\documentclass[letterpaper]{article} % DO NOT CHANGE THIS
\usepackage[submission]{aaai25}  % DO NOT CHANGE THIS
\usepackage{times}  % DO NOT CHANGE THIS
\usepackage{helvet}  % DO NOT CHANGE THIS
\usepackage{courier}  % DO NOT CHANGE THIS
\usepackage[hyphens]{url}  % DO NOT CHANGE THIS
\usepackage{graphicx} % DO NOT CHANGE THIS
\usepackage{graphicx} % DO NOT CHANGE THIS
\usepackage{natbib}  % DO NOT CHANGE THIS AND DO NOT ADD ANY OPTIONS TO IT
\usepackage{caption} % DO NOT CHANGE THIS AND DO NOT ADD ANY OPTIONS TO IT
\usepackage[dvipsnames]{xcolor}
\usepackage{algorithm}
\usepackage{algpseudocode}

\usepackage{newfloat}
\usepackage{listings}
\DeclareCaptionStyle{ruled}{labelfont=normalfont,labelsep=colon,strut=off} % DO NOT CHANGE THIS
\floatstyle{ruled}
\newfloat{listing}{tb}{lst}{}
\floatname{listing}{Listing}

\usepackage{booktabs} % 表格线条
\usepackage{multirow} % 跨行/列文本
\usepackage{graphicx} % 图形相关命令
\usepackage{amsmath}
\usepackage{amsfonts}
\usepackage{arydshln}
\usepackage{bbold}
\usepackage{bm}

\usepackage{bibentry}

\begin{document}

\title{MoE-LPR: Multilingual Extension of Large Language Models through Mixture-of-Experts with Language Priors Routing}
\author {
    % Authors
    Hao Zhou\textsuperscript{\rm 1}\equalcontrib~ Zhijun Wang\textsuperscript{\rm 1}\equalcontrib~ Shujian Huang\textsuperscript{\rm 1}\thanks{Corresponding author.}~\\ Xin Huang\textsuperscript{\rm 2}~Xue Han\textsuperscript{\rm 2}~Junlan Feng\textsuperscript{\rm 2}~Chao Deng\textsuperscript{\rm 2}~Weihua Luo\textsuperscript{\rm 3}~Jiajun Chen\textsuperscript{\rm 1}
}
\affiliations {
    % Affiliations
    \textsuperscript{\rm 1}National Key Laboratory for Novel Software Technology, Nanjing University, China\\  
    \textsuperscript{\rm 2}China Mobile Research Beijing, China; \textsuperscript{\rm 3}Alibaba Group,China\\
    \texttt{\{zhouh,wangzj\}@smail.nju.edu.cn,\{huangsj,chenjj\}@nju.edu.cn,}\\
    \texttt{\{huangxinyjy,hanxueai,fengjunlan,dengchao\}@chinamobile.com, weihua.luowh@alibaba-inc.com}
}

\maketitle

\begin{abstract}
Large Language Models (LLMs) are often English-centric due to the disproportionate distribution of languages in their pre-training data. Enhancing non-English language capabilities through post-pretraining often results in catastrophic forgetting of the ability of original languages. Previous methods either achieve good expansion with severe forgetting or slight forgetting with poor expansion, indicating the challenge of balancing language expansion while preventing forgetting.
In this paper, we propose a method called \textbf{MoE-LPR} (\textbf{M}ixture-\textbf{o}f-\textbf{E}xperts with \textbf{L}anguage \textbf{P}riors \textbf{R}outing) to alleviate this problem. MoE-LPR employs a two-stage training approach to enhance the multilingual capability. First, the model is post-pretrained into a Mixture-of-Experts (MoE) architecture by upcycling, where all the original parameters are frozen and new experts are added. In this stage, we focus improving the ability on expanded languages, without using any original language data.
Then, the model reviews the knowledge of the original languages with replay data amounting to less than 1\% of post-pretraining, where we incorporate language priors routing to better recover the abilities of the original languages.
Evaluations on multiple benchmarks show that MoE-LPR outperforms other post-pretraining methods. Freezing original parameters preserves original language knowledge while adding new experts preserves the learning ability. Reviewing with LPR enables effective utilization of multilingual knowledge within the parameters. Additionally, the MoE architecture maintains the same inference overhead while increasing total model parameters. Extensive experiments demonstrate MoE-LPR's effectiveness in improving expanded languages and preserving original language proficiency with superior scalability. Code and scripts are freely available at \url{https://github.com/zjwang21/MoE-LPR.git}.
\end{abstract}

\section{Introduction}
\begin{figure}[htb]
\centering
\includegraphics[width=0.42\textwidth]{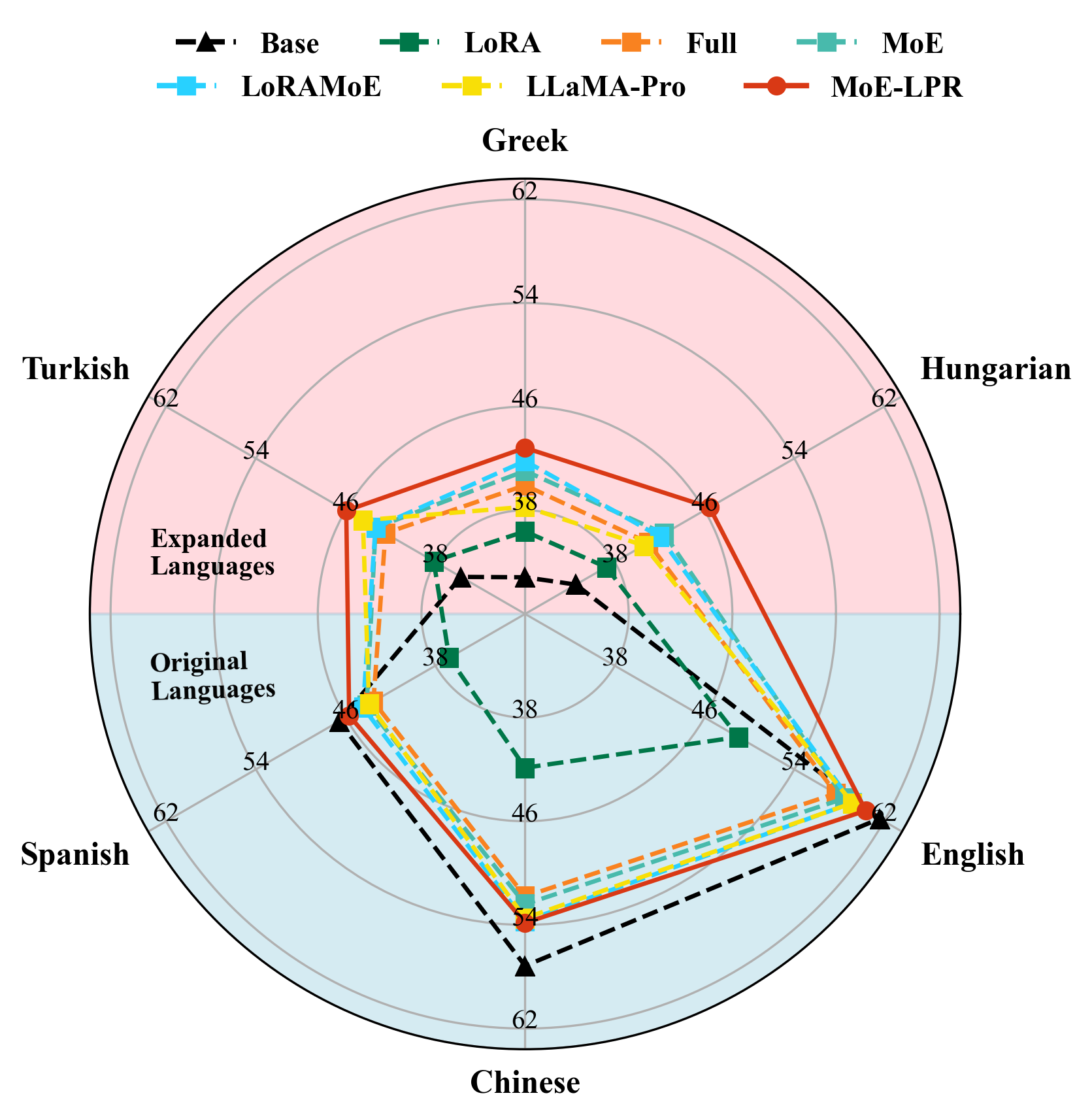}
\caption{MoE-LPR performs the best in both expanded languages and original languages. We define expanded languages as languages that the model is not very good at and we are going to enhance, and original languages as languages that the model is relatively strong in and prone to catastrophic forgetting.}
\label{xy}
\end{figure}

Large Language Models (LLMs) such as ChatGPT~\citep{chatgpt}, GPT-4~\citep{achiam2023gpt}, Llama2~\citep{touvron2023llama}, Llama3~\citep{dubey2024llama}, and Qwen~\citep{bai2023qwen} have demonstrated remarkable performance across different tasks, including multiple-choice question-answering~\citep{robinson2023leveraging}, summarization~\citep{pu2023summarization}, and reasoning~\citep{yu2023metamath}. However, many studies have highlighted a significant discrepancy between performances on English and non-English tasks~\citep{gao-etal-2024-multilingual,SeaEval}.

Pre-training a LLM with data from multiple languages may achieve better multilingual capabilities, but highly resource-intensive and often impractical given limited computational budgets. 
Consequently, current research predominantly focus on post-pretraining (also known as continue training) techniques~\citep{csaki2024sambalingo,kuulmets-etal-2024-teaching}, which carry out further multilingual pre-training on a pre-trained LLM, aiming to inject extensive language knowledge for certain language(s).
Despite its efficiency, this method significantly increases the risk of catastrophic forgetting, where the performance of LLMs in the languages they are initially good at (such as English or Chinese) may dramatically decline. As a result, improving the performance of expanded languages while maintaining the performance of existing ones becomes a critical challenge in the field. 

To prevent forgetting, existing work~\citep{dou-etal-2024-loramoe, wu-etal-2024-llama} usually retain the original parameters of the model as much as possible, and train new parameters to fit knowledge for new languages. However, less attention is paid on effectively incorporating these new and old parameters for tasks in different languages.
In this paper, we propose a novel two-stage training method called Mixture-of-Experts with Language Priors Routing~(\textbf{MoE-LPR}) that improves multilingual capability with the retention of original language proficiency.
MoE-LPR contains two stages: post-pretraining with MoE and review with LPR.

In the post-pretraining stage, we upcycle the LLM into a MoE architecture and post-pretrain the newly added parameters with a substantial amount of high-quality monolingual data, while keeping the original parameters frozen. This ensures that the original capabilities of the model are preserved while expanding its proficiency in additional languages. We also incorporate load balancing loss to unleash the model's learning potential and maintain training stability. In the review stage, we further train the router to better utilize the experts for different languages. We design LPR training to recover the model's capabilities in its original languages using replay data that amounts to less than 1\% of the post-pretraining corpus. 

As shown in Figure \ref{xy}, experiment results demonstrate that our method not only significantly improves proficiency in newly expanded languages (languages in the top half) but also substantially retains the model's capabilities in its original languages (languages in the bottom half). Moreover, our approach allows for easy upscaling for the number of model parameters while maintaining a fixed inference overhead. Our approach represents a step forward in developing LLMs that are both powerful and versatile across a wide range of languages, addressing the critical need for more inclusive and effective NLP technologies in a multilingual world. The contributions of our proposed method are as follows:
\begin{itemize}
\item Two-Stage Training Strategy: MoE-LPR employs a two-stage training strategy, with a special focus on balancing the capability of newly expanded languages and the original languages.
\item Language Priors Routing: MoE-LPR introduces the LPR mechanism to mitigate catastrophic forgetting of original languages with replay data amounting to less than 1\% of the post-pretraining corpus. LPR also exhibits excellent generalization to languages it has not been trained on.
\item Scalability: MoE-LPR is designed to easily upscale the number of model parameters without increasing the inference overhead and the risk of catastrophic forgetting, making it a cost-effective and stable solution for multilingual NLP tasks.
\end{itemize}

\section{Methodology}
\begin{figure*}[t]
\centering
\includegraphics[width=0.973\textwidth]{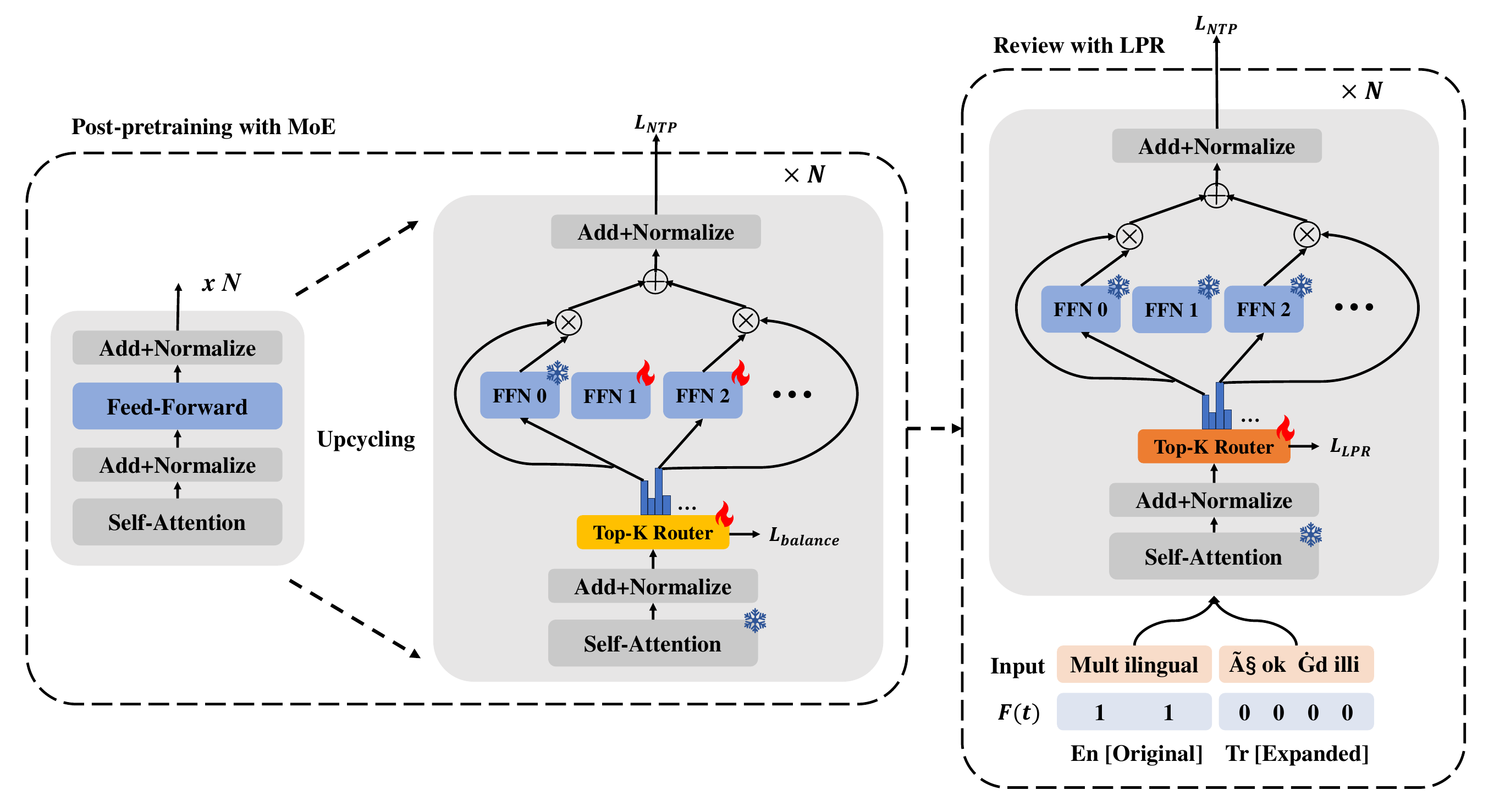}
\caption{Overall framework of our MoE-LPR. Two-stage strategy is performed to enhance the multilingual capability.}
\label{main:MoE-LPR}
\end{figure*}

Figure \ref{main:MoE-LPR} describes the overall framework of our MoE-LPR. In the post-pretraining with MoE stage, we train the new experts on a large amount of monolingual data in the expanded languages for injecting language knowledge. In the review with LPR stage, we train the router on a small amount of monolingual data in both the expanded and original languages for better utilizing the experts.

\subsection{Post-pretraining with MoE}
As shown in Figure \ref{main:MoE-LPR}, inspired by Mixtral ~\citep{jiang2024mixtral} and upcycling ~\citep{komatsuzaki2022sparse}, we upcycle the dense model to a MoE model by copying the FFN parameters and incorporating a router matrix $W_{r}\in \mathbb{R}^{h\times N}$ in each layer, where $h$ represents the token dimension and $N$ denotes the number of experts within the model. 

The router in MoE allows the model to dynamically select the most suitable experts. 
Formally, let $x \in \mathbb{R}^{h}$ be a token representation, the router score is expressed as:
\begin{equation}
    G\left ( x \right )  = \text{Softmax}\left ( x\cdot W_{r}   \right ) 
\end{equation}
\noindent where $G\left ( x \right )  \in \mathbb{R}^{N}$. After obtaining router scores, We select the index set $\mathcal{T}$ of the top-$K$ experts and combine their outputs using normalized weights from the router scores to obtain the final representation as:
\begin{equation}
    \mathcal{T} = \left \{ i| G_i\left ( x \right ) \in \text{Topk}(G\left ( x \right ), K)  \right \} 
\end{equation}
\begin{equation}
    y= \sum_{i \in \mathcal{T}}\frac{ G_{i}(x) }{\sum_{j \in \mathcal{T}} G_{j}(x)}E_{i}(x) +x
\end{equation}
\noindent where $G_{i}(x)$ and $E_{i}(x)$ represent the router score and the output of the $i$-th expert respectively, and $K$ denotes the number of activated experts.

To enhance the multilingual capability of the MoE model while preserving its performance in the original languages, we freeze the parameters of the original dense model. During post-pretraining on the expanded language corpus, we only update the parameters of the newly added experts and the router, which ensures that the core knowledge embedded in the initial model remains intact.

The model is trained with a combination of a next token prediction loss and a load balancing loss as follows.

\noindent \textbf{Next Token Prediction Loss.}
Given an expanded language corpus $D$, a batch $\mathcal{B}$ with $T$ tokens, and $N$ experts indexed by $i$ from 0 to $N-1$, where index 0 is used to denote the original dense FFN, the post-pretraining next token prediction loss is:
\begin{equation}
 L_{\text{NTP}}(\theta_{\text{new}},W_{r}) = - \sum_{i=1}^{|\mathcal{\mathcal{B}}|} \sum_{j=1}^{\left|d^{i}\right|} \log p_{\mathcal{M}}\left(d_{j}^{i} \mid d_{<j}^{i}\right)
\end{equation}
\noindent where $\mathcal{M}$ denotes the whole MoE model, $\theta_{\text{new}}$ indicates the parameters of the newly added experts and $W_{r}$ is the parameter of the router.

\noindent \textbf{Load Balancing Loss.}
We also use an expert-level load balance loss \citep{fedus2022switch} to mitigate the risk of routing collapse:
\begin{equation}
    L_{\text{balance}}(\theta_{\text{new}},W_{r}) = \sum_{i=1}^{N} f_i P_i
\end{equation}
\begin{equation}
    f_i =  \frac{N}{KT} \sum_{t \in \mathcal{B}} \mathbb{1} \{\text{Token t selects expert i}\} 
\end{equation}
\begin{equation}
    P_i = \frac{1}{T}\sum_{t \in \mathcal{B}} G_{i}(t)
\end{equation}

\noindent where $\mathbb{1}$ denotes the indicator function. We opt for a top-2 strategy by setting $K=2$ to select the two most suitable experts with normalization, intending to achieve a trade-off between inference overhead and learning capabilities. 

The final optimization objective during post-pretraining is:
\begin{equation}
\label{stage1loss}
    \underset{\theta_{\text{new}},W_{r}}{\operatorname{argmin}}  
    L_{\text{NTP}} + \alpha L_{\text{balance}}
\end{equation}
\noindent where $\alpha$ is a hyper-parameter that controls the weight of the load balancing loss.

\subsection{Review with LPR}

After post-pretraining on the expanded language corpus, the router, which has only been trained on the expanded languages but not on the original languages, may incorrectly assign experts for the original languages. This misallocation is also an important factor for catastrophic forgetting in the MoE model. 
Therefore, we design this review stage to train the model to deal with both original and expanded languages. 

As the router is the main source of the problem, we only update the parameters of the router and freeze the other parts of the model. Because the number of router parameters accounts for a negligible proportion, this stage could be efficient and requires very little computational resource and training data.

In fact, the amount of original language data used in our review stage, is less than 1\% of the post-pretraining corpus. In comparison, traditional replay strategy~\citep{ibrahim2024simple} incorporates data from original languages into the post-pretraining stage, which usually requires a much larger amount (25\%).

\noindent \textbf{LPR Loss.} Intuitively, the routing could be led by language priors: all the original language tokens should be routed to the originally frozen expert (i.e. expert 0 in this case), making the model work exactly the same as before the expansion. 
Therefore, we design a LPR loss to be a Cross-Entropy loss for the tokens from the original languages, forcing the top-1 selection of these tokens to be expert 0, where the top-1 selection refers to the expert selection with the highest routing score. 

Formally, considering original language tokens set $D_{\text{original}}$ and the indicator function $\mathbf{F}(t)$ :
\begin{equation}
\mathbf{F}(t) = \begin{cases}
1 & \text{if } t \in D_{\text{original}}, \\
0 & \text{if } t \notin D_{\text{original}}.
\end{cases}
\end{equation}
\noindent The LPR loss is defined as:
\begin{equation}
\label{lpr}
    L_{\text{LPR}}(W_r) = -\sum_{t \in \mathcal{B}} \mathbf{F}(t)\log G_{0}(t)
\end{equation}
\noindent where index 0 denotes the originally frozen expert. 

In practice, when training with LPR loss, we remove the load balancing loss in Eq.~\eqref{stage1loss}. The final optimization objective for the review stage is:
\begin{equation}
    \underset{W_{r}}{\operatorname{argmin}}
    L_{\text{NTP}} + \gamma L_{\text{LPR}}
\end{equation}
\noindent where $\gamma$ is a hyper-parameter that controls the weight of the LPR loss.

\section{Experiments}
\subsection{Experiment Setup}
Given the focus on multilingual capability enhancement, we introduce the language selection first. Then follow the training details, several baselines, and the evaluation details.

\paragraph{Model and Languages}
We choose Qwen-1.5~\footnote{Qwen-1.5 has a powerful multilingual tokenizer that produces shorter sequences in expanded languages, which means we don't have to worry about vocabulary expansion.} as our base model. The 1.8B version of the Qwen-1.5 series is selected for its lower computation overhead and ease of upcycling.
For our study, we choose three low-resource languages as the expanded languages where Qwen-1.5-1.8B performs poorly as shown in Figure \ref{xy}: Greek (El), Hungarian (Hu), and Turkish (Tr). Additionally, we select three high-resource languages as the original languages to observe the catastrophic forgetting phenomenon: English (En), Chinese (Zh), and Spanish (Es).

\paragraph{Details of Post-pretraining}
We construct a dataset focusing on the three expanded languages by sampling 8 billion tokens from the monolingual data of each language in CulturalX~\citep{nguyen-etal-2024-culturax-cleaned}, a substantial multilingual dataset with 6.3 trillion tokens in 167 languages. Our base model, Qwen-1.5-1.8B, is upcycled into the MoE structure with 5 newly added FFN (6 experts in total). We post-pretrain this model with the 24 billion tokens, marking only the new experts and the router as trainable. The training setup includes a batch size of 512, a sequence length of 1024, a learning rate of 5e-5, and a cosine learning rate scheduler. We incorporate the load balancing loss with a weight of 0.01 and utilize bf16 mixed precision and flash attention~\citep{dao2022flashattention} to speed up the training process.

Our experiments are conducted on 8 A800 GPUs, involving 45856 steps, totaling approximately 848 A800 GPU hours.

\paragraph{Details of Review}
We randomly sample 50K documents for each original language and 100K documents for each expanded language. The English data are sampled from Slimpajam~\citep{cerebras2023slimpajama}, the Chinese data from SkyPile-150B~\citep{wei2023skywork}, and the Spanish data from CulturalX~\citep{nguyen-etal-2024-culturax-cleaned}.
The number of tokens in original languages is 0.138B, accounting for less than 1\% of the post-pretraining data (24B).
As for the three expanded languages, we sample from the post-pretraining dataset. We concatenate these data for the review stage training. We employ a batch size of 512, a sequence length of 512, a learning rate of 5e-5, and a cosine learning rate scheduler. The load balancing loss is removed and the LPR loss is added as introduced in Eq.~\eqref{lpr} with a weight of 0.1. Only the router parameters are trainable. Bf16 mixed precision and flash attention~\citep{dao2022flashattention} mechanism is used for training.

\paragraph{Baselines}
We conducted experiments on several existing baseline methods trained on the same data, including the small amount of replay data, to ensure that our approach is competitive and effective.

\begin{itemize}
    \item \textbf{Full Fine-tuning}: Fine-tune all parameters directly on the dense model.
    \item \textbf{LoRA}~\citep{hu2021lora}: The LoRA targets include all linear modules. We set the LoRA rank to 8.
 
    \item \textbf{MoE}: The same settings as MoE-LPR except for training all the parameters only in one post-pretraining stage.

    \item \textbf{LoRAMoE}~\citep{dou-etal-2024-loramoe}: A novel framework combines multiple LoRAs with a router network to effectively learn new knowledge while avoiding catastrophic forgetting. The router selects all LoRAs for each token. We set the number of LoRAs as 8 and a LoRA rank of 180 to match the same inference overhead.
    
    \item \textbf{LLaMA-Pro}~\citep{wu-etal-2024-llama}: A method is considered where a dense LLM periodically duplicates and inserts new transformer blocks at fixed layer intervals. During post-pretraining, only these newly added transformer blocks are trained to acquire new knowledge while preserving the original knowledge. We add 12 new layers because this is the best setting in our experiments.
\end{itemize}

\begin{table*}[t]
\centering
{
\begin{tabular}{lcccccccc}
\toprule
{\bf Model} &{\bf $\bm{n_{\textbf{params}}}$} &{\bf $\bm{n_{\textbf{act-params}}}$} &{\bf ARC} &{\bf MMLU} &{\bf HellaSwag} &{\bf Belebele} &{\bf Flores} &{\bf Avg.} \\
\midrule
\bf{Expanded Languages} &&&&&& \\
\midrule
Qwen1.5-1.8B &1.8B &1.8B &23.13 &30.97 &29.15 &33.15 &55.40 &34.36\\
\hdashline
LoRA~\scriptsize{~\citep{hu2021lora}} &1.8B &1.8B &23.89 &29.30 &29.78 &26.93 &55.19 &33.02\\
Full Fine-tuning &1.8B &1.8B &25.98 &33.18 &35.28 &\underline{33.70} &77.48 &41.12\\
LLaMA-Pro~\scriptsize{~\citep{wu-etal-2024-llama}}  &2.4B &2.4B  &24.35  &34.02  &33.85  &31.52  &\underline{81.76}  &41.10 \\
MoE &5.8B &2.6B &26.43 &\bf35.07 &37.01 &32.74 &80.01 &42.25\\
LoRAMoE~\scriptsize{~\citep{dou-etal-2024-loramoe}} &2.6B &2.6B &\underline{26.63} &\underline{34.17} &\underline{37.17} &32.81 &81.09 &42.37 \\
MoE-LPR&5.8B &2.6B &\bf28.43 &34.10 &\bf41.06 &\bf39.93 &\bf81.83 &\bf45.07 \\
\midrule
\bf{Original Languages} &&&&&& \\
\midrule
Qwen1.5-1.8B &1.8B &1.8B &33.48 &47.55 &49.82 &56.52 &82.50 &53.97\\
\hdashline
LoRA~\scriptsize{~\citep{hu2021lora}} &1.8B &1.8B &28.33 &37.42 &41.48 &39.45 &75.49 &44.43\\
Full Fine-tuning &1.8B &1.8B &31.72 &43.51 &47.38 &45.26 &80.77 &49.73\\
LLaMA-Pro~\scriptsize{~\citep{wu-etal-2024-llama}}  &2.4B &2.4B &31.77 &44.06 &48.36 &\underline{48.78} &81.97 &50.99\\
MoE &5.8B &2.6B &\underline{32.51} &44.16 &48.54 &45.37 &81.63 &50.44 \\
LoRAMoE~\scriptsize{~\citep{dou-etal-2024-loramoe}} &2.6B &2.6B &32.43 &\bf45.41 &\underline{48.61} &47.74 &\underline{82.03} &51.24\\
MoE-LPR &5.8B &2.6B &\bf32.71 &\underline{44.62} &\bf49.12 &\bf51.81 &\bf82.36 &\bf52.12 \\
\bottomrule
\end{tabular}}
\caption{Evaluation results in expanded and original languages. $\bm{n_{\textbf{params}}}$ is the total number of model parameters, $\bm{n_{\textbf{act-params}}}$ is the number of activated model parameters per token. The best and second-best results are marked in \textbf{bold} and \underline{underlined} fonts.}
\label{tab:main}
\end{table*}

\paragraph{Evaluation Details}
We evaluate our method on several benchmarks including multiple-choice tasks and generation tasks. Examining the model's multilingual capabilities from multiple perspectives.
\begin{itemize}
    \item \textbf{ARC-Challenge (25-shot)}~\citep{clark2018think}: A benchmark for evaluating comprehension and reasoning across diverse academic fields.

    \item \textbf{MMLU (5-shot)}~\citep{hendrycks2020measuring}: A multiple-choice dataset testing general knowledge and problem-solving across various subjects.

    \item \textbf{HellaSwag (10-shot)}~\citep{zellers2019hellaswag}: A dataset with 70k questions for studying grounded commonsense inference.
    
    \item \textbf{Belebele (5-shot)}~\citep{bandarkar2023belebele}: A machine reading comprehension dataset covering 122 language variants.

    \item \textbf{FLORES-101 (8-shot)}~\citep{goyal2022flores}: A parallel corpus for evaluating multilingual translation capabilities. We report the performance evaluated by COMET~\citep{rei2022comet}~\footnote{We use the \textrm{wmt22-comet-da} version.}
\end{itemize}
We mainly follow Okapi~\cite{lai2023okapi} to evaluate the multilingual versions of ARC-Challenge, MMLU and HellaSwag, which are translated from the original English version using GPT-3.5-turbo or DeepL. More details about the source are reported in the technical appendix C.

\subsection{Experiment Results}
Table \ref{tab:main} presents the performance of various methods across different benchmarks for both expanded and original languages. We report here the performance of the best setting of all baselines. With the additional small amount of replay data, full fine-tuning outperforms LoRA in preventing catastrophic forgetting but still drops about 4 points in original languages. 
Full fine-tuning can recover to 92.1\% performance in original languages with replay data amounting to less than 1\% of the post-pretraining data. \citet{ibrahim2024simple} demonstrates that training new languages suffers from dramatic distribution shifts. Only when using more than 25\% replay data can the model recover to more than 95.7\% performance, indicating that significant language shifts in post-pretraining data require more replay data and computational overhead. However, our MoE-LPR can recover to 96.6\% performance (52.12/53.97) with less than 1\% replay data. 

LoRA performs poorly in expanded languages due to the excessive data in the post-pretraining stage. We also experiment with LoRA at rank=64 to achieve comparable effects in expanded languages, but this results in worse catastrophic forgetting, as shown in the technical appendix B.

LLaMA-Pro demonstrates a strong ability to retain knowledge, but its performance in expanded languages is only comparable to full fine-tuning, with the drawback of higher inference overhead. LoRAMoE performs better than other baselines in both expanded and original languages. Our proposed method, MoE-LPR, surpasses LoRAMoE by 2.7 points in expanded languages and by 0.88 points in original languages on average.
While adding more new parameters, the inference overhead of LLaMA-Pro and LoRAMoE increases accordingly, while that of MoE-LPR does not. More details about scaling will be discussed in the following sections.

The results also demonstrate that MoE underperforms our MoE-LPR both in expanded and original languages, which implies that freezing all the original parameters will not limit the model's learning ability. In contrast, the frozen parameters contribute a robust basic capabilities of the model during post-pretraining, resulting in significant performance improvement. Further details on each benchmark are provided in the technical appendix A.

\section{Abalation \& Analysis}
\subsection{Review with LPR}

\subsubsection{Performance Gain from Review \& EC} \label{lprablation}
\begin{table}[htb]
\centering
\setlength{\tabcolsep}{1mm}
{
\begin{tabular}{lccc}
\toprule
{\bf Model} &{\bf Expanded} &{\bf Original} &{\bf Avg.} \\
\midrule
Qwen1.5-1.8B &34.36 &53.97 &44.17\\
\hdashline
LoRAMoE &42.37 &51.24 &46.81\\
MoE-LPR w/o EC &38.37 &49.28 &43.83\\
MoE-LPR w/o Review &45.04 &47.14 &46.09\\
MoE-LPR w/o LPR &\bf45.13 &51.32 &48.23\\
MoE-LPR &45.07 &\bf52.12 &\bf48.60\\
\bottomrule
\end{tabular}}
\caption{Evaluation average results with different settings. ``w/o EC'' means without expert-copy, corresponding to randomly initialize the new experts when upcycling.}
\label{tab:lprgain}
\end{table}

The review with LPR stage is proposed to recover the capabilities of the original languages. As shown in Table \ref{tab:lprgain}, without the review stage, MoE-LPR exhibits severe catastrophic forgetting. However, after review training, the performance in original languages improves substantially, by about 5 points on average, while not harming the performance in expanded languages. Furthermore, the performance in original languages drops without the LPR loss, indicating that the LPR mechanism pushes this ability closer to its upper bound. These results show that the review stage allows the model to learn how to handle both new and old languages.

We also conduct experiment without the Expert-Copy, which means that the parameters of new experts are randomly initialized but not copied from the original FFN. As shown in Table \ref{tab:lprgain}, performance in original languages does not suffer a serious decrease, but performance in expanded languages shows a significant decrease. Results imply that copying the original FFN to construct new experts is important to the learning of expanded language knowledge.

\begin{figure}[htb]
\centering
\includegraphics[width=0.48\textwidth]{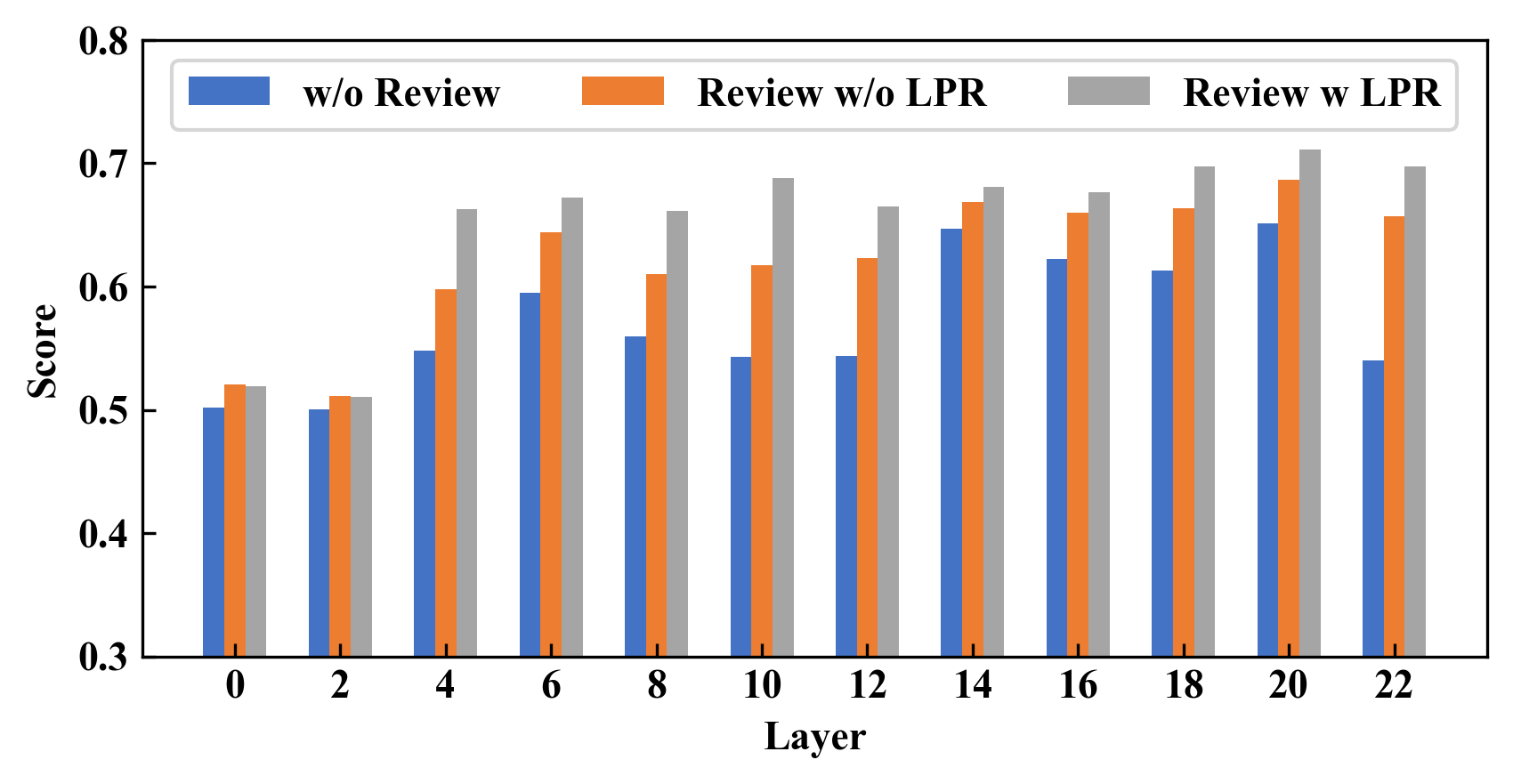}
\caption{Router scores of the frozen expert for English (original language) tokens in the Belebele benchmark.}
\label{figure:untrained_score}
\end{figure}

\subsubsection{Routing Scheme for Different Languages}
In this section, we examine whether the review stage works properly. As shown in Figure \ref{figure:untrained_score}, the router scores of the frozen expert on original language tokens show obvious improvement with the review stage. In addition, without the LPR loss, the router scores demonstrate a significant drop.
The router scores of the frozen expert on expanded language tokens almost remain unchanged, as shown in the technical appendix D. In the review stage, we optimize the model with only the next token prediction loss for expanded languages. The results show that the next token prediction loss effectively prevents expanded languages from being influenced by the language priors of original languages.
These observations indicate that the review stage is functioning correctly, biasing the routing scheme of original language tokens toward the frozen expert.

\begin{figure}[htb]
\centering
\includegraphics[width=0.48\textwidth]{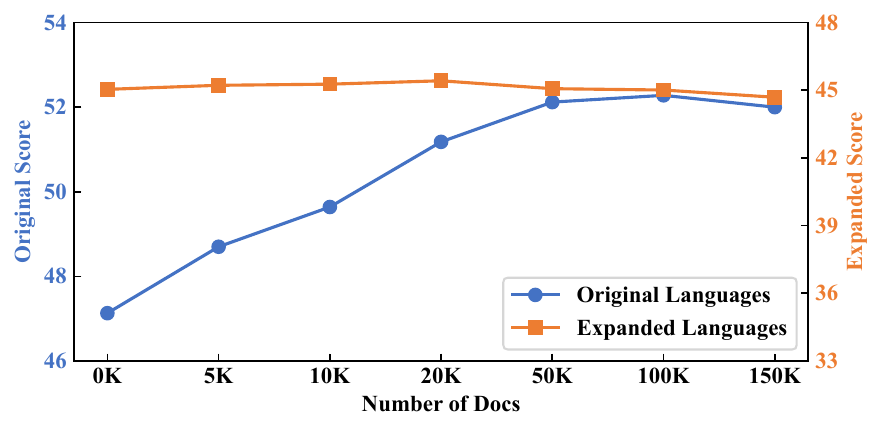}
\caption{Average scores in expanded and original languages with varying numbers of documents for review.}
\label{figure:lprscale}
\end{figure}

\subsubsection{How much Data is Enough for Review}
In this section, we experiment with varying numbers of original language documents in the review stage, ranging from 0K to 150K, while maintaining the 1:2 mix of original and expanded languages. As shown in Figure \ref{figure:lprscale}, the original language performance continues to improve significantly while the expanded language performance continues to decrease slightly. 
After 50k, the original language performance improvement starts to become slow. Therefore, considering both training cost and effects, we choose 50K as the best data size in this experiment, which amounts to less than 1\% of the post-pretraining corpus.
Using 50K results in a 4.98 points performance boost in the original languages while almost maintaining the performance in the expanded languages. These results indicate that a small amount of replay data is sufficient for the model to review its original languages.

\begin{figure}[htb]
\centering
\includegraphics[width=0.48\textwidth]{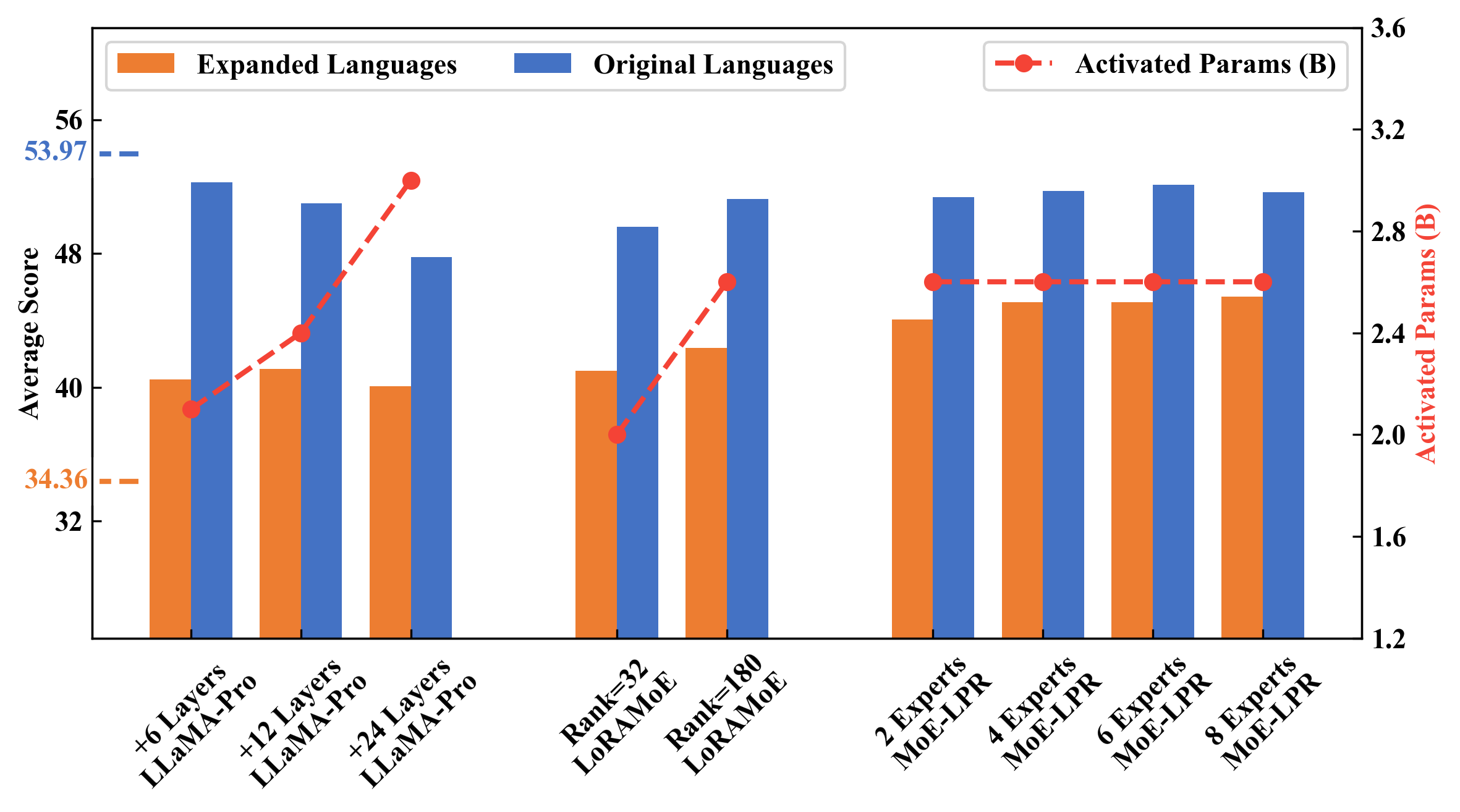}
\caption{Average scores in expanded and original languages with different model settings. ``34.36'' and ``53.97'' refer to the expanded and original language performance of the base model respectively.}
\label{figure:scale}
\end{figure}

\subsection{Scaling Law} \label{scaling}
We compare the performance of LLaMA-Pro with different numbers of extending layers, LoRAMoE with different ranks and MoE-LPR with different numbers of experts. All the models are trained on the 24 billion tokens dataset in the three expanded languages.

Figure \ref{figure:scale} demonstrates the superior scalability of MoE-LPR. For expanded languages, adding 12 layers to LLaMA-Pro improves performance more than adding 6 layers, but adding 24 layers, matching the base model's layer count, results in a performance drop. Increasing the rank of LoRAMoE from 32 to 180 shows significant improvements. MoE-LPR consistently outperforms these configurations as more experts are added, even with just 2 experts, maintaining a significant advantage over LLaMA-Pro and LoRAMoE.
For original languages, LLaMA-Pro suffers from catastrophic forgetting, worsening with more layers. Adding 24 layers even performs worse than full fine-tuning. Although LoRAMoE's catastrophic forgetting does not worsen with increased parameters, it still underperforms MoE-LPR. Even with 8 experts and a 7B parameter size, MoE-LPR can still greatly mitigate catastrophic forgetting.

Unlike LLaMA-Pro and LoRAMoE, whose activated parameters per token increase linearly with more parameters, adding experts to MoE-LPR does not increase the inference overhead. This improves performance in expanded languages while maintaining stable levels of catastrophic forgetting. MoE-LPR demonstrates superior scalability. Details of each model in the scaling experiments are reported in the technical appendix A.

\begin{table}[ht]
\centering
{
\begin{tabular}{lccc}
\toprule
{\bf Model} &{\bf Exp.} &{\bf Ori. ID} &{\bf Ori. OOD}\\
\midrule
Qwen1.5-1.8B &34.36 &53.97 &46.35 \\
\hdashline
Full Fine-tuning &41.12 &49.73 &42.46 \\
LLaMA-Pro &41.10 &50.99 &42.93 \\
LoRAMoE &42.37 &51.24 &43.41 \\
MoE-LPR w/o LPR &45.13 &51.32 &44.22 \\
MoE-LPR One-Stage &\bf45.38 &51.90 &43.71 \\
MoE-LPR &45.07 &\bf52.12 &\bf45.25 \\
\bottomrule
\end{tabular}}
\caption{Evaluation results in French and Portuguese.}
\label{tab:generalization}
\end{table}

\subsection{Language Generalization}
In the review stage, we only use documents of three of the original languages. We conduct evaluations on two additional high-resource languages that the base model is good at relatively: French and Portuguese to examine the generalization of MoE-LPR when preventing catastrophic forgetting. We name them out-of-domain original languages because the review stage training does not contain tokens in these two languages. Table \ref{tab:generalization} demonstrates that MoE-LPR successfully generalizes its catastrophic forgetting prevention effect to these languages. Despite the router not being trained on French and Portuguese tokens, our LPR mechanism minimizes the performance gap from the base model for these languages, outperforming other post-pretraining methods. This demonstrates MoE-LPR's excellent language generalization in preventing catastrophic forgetting.

We also try to move the LPR loss and the small amount of replay data to the post-pretraining stage. As shown in Table \ref{tab:generalization}, MoE-LPR One-Stage shows comparable performance to the two-stage strategy. However, it demonstrates worse language generalization, which showcases a 1.54 points performance drop in the out-of-domain original languages. Therefore, we choose the two-stage strategy as a better proposal.

\section{Related Work}

\subsection{Mixture of Experts}

Recent studies~\citep{kaplan2020scaling,hoffmann2022empirical} have shown a strong correlation between the number of parameters in a model and its capabilities. When the number of parameters is large, the model demonstrates emergent abilities~\citep{52065}. Traditional dense models require the activation of all parameters for a given input, significantly increasing computational overhead. Distinct from conventional dense models, Mixture of Experts~(MoE) achieves computational feasibility and expanded model capacity by utilizing a router that selectively activates a limited number of experts for each input. There are several works, such as Switch-transformer~\citep{fedus2022switch}, ST-MoE~\citep{zoph2022st}, Glam~\citep{du2022glam} , attempts to train an MoE model from scratch. These works have demonstrated that MoE models can achieve significantly lower loss and performance gains compared to dense models with the same activated parameters and require less energy consumption compared to dense models with the same total parameters. However, considering the huge computational budget, \citet{komatsuzaki2022sparse} indicates that a sparse MoE model could be initialized from dense models. In the era of LLMs, numerous MoE works have been developed. For instance, Mixtral~\citep{mixtral} adds experts to each layer, increasing the total parameter count to 141B. DeepSeek~\citep{deepseekai2024deepseekv2} utilizes shared experts, enabling the model to select experts more effectively. Snowflake Arctic~\citep{SnowflakeArctic}incorporates many fine-grained experts, enhancing the diversity of expert selection. \citet{chen2023octavius,dou-etal-2024-loramoe,zadouri2023pushing} combines MoE with LoRA, resulting in more effective training and alleviating data conflict issues.

The most relevant work to us is Lifelong-MoE~\citep{chen2023lifelong}, which effectively expands the number of experts during lifelong learning and introduces a regularization to avoid catastrophic forgetting. However, we employ a different freezing method and a two-stage training framework, significantly alleviating catastrophic forgetting and gaining a promising performance in expanded languages.

\subsection{LLM for Multilingual}
Post-pretraining on a massive multilingual corpus is an effective way to improve the multilingual abilities of LLMs. \citet{alves2024tower} and \citet{xu2024a} highlight monolingual data's importance in post-pretraining. Notably, \citet{xu2024a} demonstrates that with fixed computational resources, allocating more to monolingual data rather than translation data better improves a model's translation performance, allowing large models to achieve translation abilities comparable to traditional supervised models NLLB~\citep{costa2022no}. \citet{blevins2024breaking} have explored using the Branch Then Merge~(BTM;\citet{gururangan2023scaling}), where separate models are trained independently for different languages and then merged, partially overcoming the challenges of the multilingual curse~\citep{wu2020all}. \citet{geng2024not} employs the LoRA~\citep{hu2021lora} architecture to help migrate a chat LLM to the target language while preserving its chat capabilities.
 
\section{Conclusion}
In this paper, we propose MoE-LPR, a scalable post-pretraining method that effectively expands languages and prevents catastrophic forgetting using the Mixture-of-Experts architecture. Expanding new languages often encounters severe catastrophic forgetting due to significant distribution changes, and the challenge lies in balancing old and new languages. Through two-stage training, MoE-LPR addresses this with efficient parameter assignment and balanced routing.
The post-pretraining stage enables the model to have a strong enough learning ability and steadily enhances the capabilities of the expanded languages.
The review stage brings a performance boost to the original languages without harming the performance in expanded languages.
Our two-stage training achieves both expansion and prevention of forgetting effects well.
Additionally, MoE-LPR shows better scalability and generalization than SOTA methods. Overall, MoE-LPR is an effective and scalable approach for expanding new languages during the post-pretraining stage.

\bibliography{aaai25}

\appendix
\section{A}
Experiment results in detail for each benchmark and language are shown in Table \ref{tab:allres} and Table \ref{tab:Flores}. The performance of the best configuration for each method is presented. The results demonstrate that MoE-LPR outperforms the comparison methods on most benchmarks, especially HellaSwag and Belebele.

The results of the scaling experiment are detailed in Table \ref{tab:scale}.
With the addition of more new experts, the performance of MoE-LPR in expanded languages continues to increase while demonstrating stable effects in preventing catastrophic forgetting of original languages. Meanwhile, the inference overhead per token of MoE-LPR remains unchanged, showing more powerful scalability than other methods.

\section{B}

Hyper-parameters for different baselines.

\begin{table}[ht]
\setlength{\tabcolsep}{1mm}
\centering
{
\begin{tabular}{lcccc}
\toprule
{\bf Model} &{\bf LR} &{\bf Rank} &{\bf Expanded} &{\bf Original} \\
\midrule
Qwen1.5-1.8B &N/A &N/A &34.36 &53.97 \\
Full Fine-tuning &1e-5 &N/A &41.23 &47.94 \\
Full Fine-tuning &5e-6 &N/A &41.12 &49.73 \\
LoRA &2e-4 &8 &33.02 &44.43 \\
LoRA &2e-4 &64 &37.25 &40.71 \\
\bottomrule
\end{tabular}}
\caption{Hyperparameter configurations and performances of different baselines }
\label{tab:hp}
\end{table}

\subsection{Full Fine-tuning of dense and MoE}
In our experiments, the hyperparameters for full fine-tuning remain the same as those in MoE-LPR, except for the learning rate. Our experimental results demonstrate that full fine-tuning requires a smaller learning rate. As shown in Table \ref{tab:hp}, we experimented with a series of learning rates and determined that 5e-6 is the best configuration for full fine-tuning of the dense model. Meanwhile, 5e-6 is also the best configuration for full fine-tuning of MoE. When using a larger learning rate, the final checkpoint demonstrates random selection on several multiple-choice benchmarks, indicating that a drastic parameter distribution shift has occurred, causing the model to lose basic capabilities.

\subsection{LoRA}
We conducted experiments for LoRA with ranks of 8 and 64, with the aim of achieving excellent catastrophic forgetting prevention and expansion effects. The results in Table \ref{tab:hp} demonstrate that a rank of 64 does show better performance in expanded languages compared to a rank of 8 but still underperforms full fine-tuning. Besides, LoRA with a rank of 64 encounters a serious catastrophic forgetting phenomenon. Results imply that continuing to improve Lora's rank will hardly make its expanded language effects exceed full fine-tuning, and will produce more serious catastrophic forgetting. Therefore, in our paper, we report the performance of rank 8. 

\subsection{LLaMA-Pro \& LoRAMoE}
For LLaMA-Pro, training details remain the same as in MoE-LPR, except for a learning rate of 2e-4 as referenced in the original paper. 
For LoRAMoE, the learning rate is 2e-4, the LoRA alpha is 32, and the dropout rate is 0.05. We set the number of LoRA as 8, a blc-alpha of 0.1, and a blc-weight of 0.1. In our experiments, we attempt to increase the inference overhead of LoRAMoE to match that of MoE-LPR. When increasing the number of LoRA, the training costs become too high since LoRAMoE activates all the LoRAs for each token, resulting in more FLOPs within the MoE architecture. Therefore, we achieve this by increasing the LoRA rank.

\begin{figure}[htb]
\centering
\includegraphics[width=0.45\textwidth]{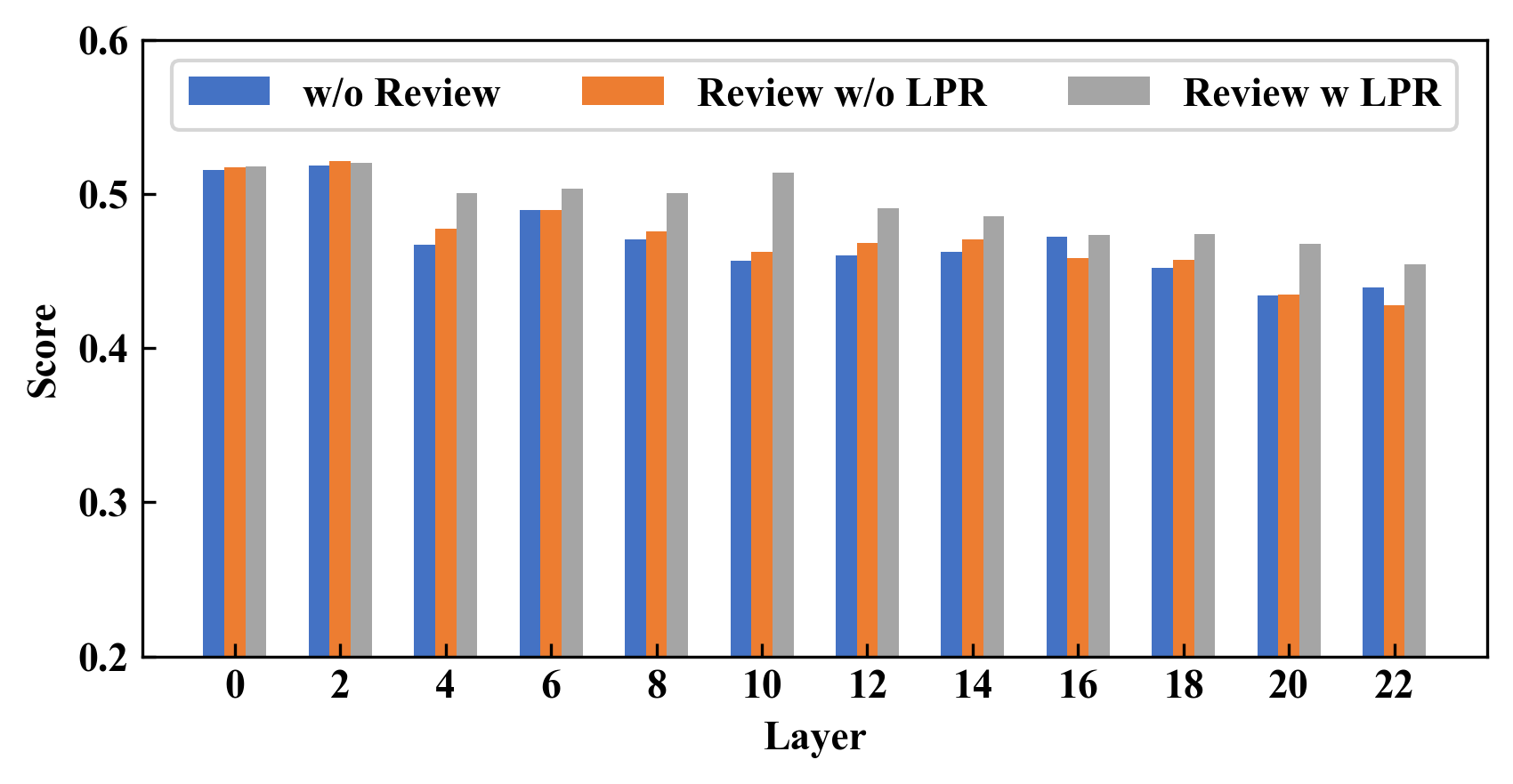}
\caption{Router scores of the frozen expert for Hungarian (expanded language) tokens in the Belebele benchmark.}
\label{figure:trained_score}
\end{figure}

\section{C}
The details of our evaluation benchmarks.
For Chinese, Spanish, Hungarian, French and Portuguese, we use the translated version of ARC-Challenge, MMLU and HellaSwag following Okapi~\citep{lai2023okapi} except for CMMLU~\citep{li2023cmmlu} in Chinese. 
For Turkish, we follow~\citep{openllm-Turkish-leaderboard} to evaluate these three benchmarks.
For Greek, we use the translated version provided by ILSP~\footnote{\url{https://huggingface.co/ilsp}}. 
The evaluation is performed through the Eleuther AI Language Model
Evaluation Harness~\footnote{\url{https://github.com/EleutherAI/lm-evaluation-harness}} except for flores, which is evaluated through script written by ourselves.

\section{D}
Figure \ref{figure:trained_score} demonstrates the router scores of the frozen expert for Hungarian (expanded language) tokens. The review stage does not significantly alter the expanded languages, ensuring their performance remained stable. Among the side effects during this phase, LPR loss was the most prominent. However, this did not significantly affect the abilities of expanded languages. Furthermore, LPR loss effectively prevents catastrophic forgetting and enhances generalization.

\begin{table*}[htb]
\centering
{
\begin{tabular}{lcccccccc}
\toprule
\multirow{2}{*}{\bf Model} & \multicolumn{3}{c}{\bf Original Languages} & \multirow{2}{*}{\bf Avg} & \multicolumn{3}{c}{\bf Expanded Languages} & \multirow{2}{*}{\bf Avg} \\
\cmidrule(lr){2-4} \cmidrule(lr){6-8}
& {\bf En} &{\bf Es} &\bf Zh & &\bf El &\bf Hu &\bf Tr & \\ 
\midrule
{\bf ARC-Challenge (25-shot)} &&&&&&&&\\
\midrule
Qwen1.5-1.8B &38.48 &27.78 &34.19 &33.48 &23.03 &23.29 &23.06 &23.13\\
\hdashline
LoRA  &31.48 &23.93 &29.57 &28.33 &24.40 &23.72 &23.57 &23.89\\
Full Fine-tuning &36.86 &26.75 &31.54 &31.72 &25.60 &26.11 &26.22 &25.98\\
LLaMA-Pro &36.01 &26.15 &33.16 &31.77 &22.43 &23.54 &27.07 &24.35 \\
MoE &37.37 &28.03 &32.14 &32.51 &25.26 &26.88 &27.16 &26.43\\
LoRAMoE  &36.18 &28.12 &32.99 &32.43 &26.46 &26.54 &26.90 &26.63\\
MoE-LPR &37.12 &27.18 &33.85 &32.71 &27.05 &29.79 &28.44 &28.43 \\
\midrule
{\bf MMLU (5-shot)} &&&&&&&&\\
\midrule
Qwen1.5-1.8B &46.36 &38.62 &57.68 &47.55 &28.98 &32.08 &31.86 &30.97\\
\hdashline
LoRA&38.93 &32.41 &40.91 &37.42 &27.99 &29.87 &30.04 &29.30\\
Full Fine-tuning &42.84 &36.34 &51.36 &43.51 &31.68 &33.88 &34.00 &33.18\\
LLaMA-Pro &44.48 &36.40 &51.31 &44.06 &31.98 &34.94 &35.15 &34.02 \\
MoE &43.59 &36.65 &52.24 &44.16 &33.41 &36.40 &35.39 &35.07\\
LoRAMoE &45.16 &37.15 &53.93 &45.41 &32.27 &36.07 &34.16 &34.17\\
MoE-LPR &44.82 &37.65 &51.38 &44.62 &31.11 &36.15 &35.05 &34.10\\
\midrule
{\bf HellaSwag (10-shot)} &&&&&&&&\\
\midrule
Qwen1.5-1.8B &62.05 &39.28 &48.12 &49.82 &28.97 &29.70 &28.77 &29.15 \\
\hdashline
LoRA  &50.44 &33.78 &40.22 &41.48 &29.23 &29.97 &30.13 &29.78\\
Full Fine-tuning &57.86 &38.19 &46.10 &47.38 &34.00 &35.98 &35.86 &35.28 \\
LLaMA-Pro &59.55 &38.80 &46.73 &48.36 &28.22 &34.33 &39.01 &33.85 \\
MoE &59.50 &39.27 &46.84 &48.54 &35.75 &37.66 &37.64 &37.01\\
LoRAMoE  &59.71 &39.08 &47.03 &48.61 &36.19 &37.93 &37.38 &37.17 \\
MoE-LPR &61.25 &39.26 &46.84 &49.12 &39.76 &41.65 &41.77 &41.06\\
\midrule
{\bf Belebele (5-shot)} &&&&&&&&\\
\midrule
Qwen1.5-1.8B &61.56 &47.33 &60.67 &56.52 &32.67 &31.89 &34.89 &33.15 \\
\hdashline
LoRA  &44.22 &32.56 &41.56 &39.45 &27.00 &24.00 &29.78 &26.93\\
Full Fine-tuning &51.00 &38.33 &46.44 &45.26 &31.89 &31.00 &38.22 &33.70 \\
LLaMA-Pro  &55.89 &38.56 &51.89 &48.78 &29.22 &26.67 &38.67 &31.52 \\
MoE &51.89 &37.44 &46.78 &45.37 &31.11 &30.33 &36.78 &32.74\\
LoRAMoE  &54.78 &38.33 &50.11 &47.74 &33.33 &27.78 &37.33 &32.81 \\
MoE-LPR &58.67 &44.33 &52.44 &51.81 &34.11 &42.22 &43.44 &39.93\\
\bottomrule
\end{tabular}}
\caption{Evaluation results on the multiple-choice benchmarks for each language.}
\label{tab:allres}
\end{table*}

\begin{table*}[!htb]
\centering
{
\begin{tabular}{lcccccccccccc}
\toprule
\multirow{2}{*}{\bf Model} & \multicolumn{4}{c}{\bf Original Languages} & \multirow{2}{*}{\bf Avg} & \multicolumn{6}{c}{\bf Expanded Languages} & \multirow{2}{*}{\bf Avg} \\
\cmidrule(lr){2-5} \cmidrule(lr){7-12}
& {\bf en-zh} &{\bf zh-en} & {\bf en-es} &{\bf es-en} &\bf  & {\bf en-el} &{\bf el-en} & {\bf en-hu} &{\bf hu-en} & {\bf en-tr} &{\bf tr-en} & \\ 
\midrule
Qwen1.5-1.8B &86.07 &84.29 &76.16 &83.50 &82.50 &40.28 &60.75 &41.05 &70.32 &50.86 &69.18 &55.40\\
\hdashline
LoRA  &80.87 &80.31 &63.97 &76.83 &75.49 &46.38 &57.70 &45.05 &65.42 &51.55 &65.03 &55.19\\
Full Fine-tuning &84.24 &82.91 &73.70 &82.25 &80.77 &72.79 &80.21 &74.99 &80.95 &76.06 &79.87 &77.48\\
LLaMA-Pro &84.86 &83.74 &76.18 &83.09 &81.97 &77.26 &81.76 &82.80 &83.96 &81.73 &83.05 &81.76\\
MoE &84.74 &83.34 &75.50 &82.95 &81.63 &77.53 &81.82 &78.88 &82.53 &77.59 &81.74 &80.01\\
LoRAMoE  &85.29 &83.76 &76.22 &82.86 &82.03 &79.31 &82.29 &80.07 &83.04 &79.31 &82.51 &81.09 \\
MoE-LPR &85.11 &84.36 &76.49 &83.48 &82.36 &83.37 &80.77 &82.05 &82.82 &80.62 &81.39 &81.83\\
\bottomrule
\end{tabular}}
\caption{Evaluation results on the Flores benchmarks (8-shot).}
\label{tab:Flores}
\end{table*}

\begin{table*}[htb]
\centering
{
\begin{tabular}{lcccccccc}
\toprule
\midrule
{\bf Model}&{\bf $\bm{n_{\textbf{params}}}$} &{\bf $\bm{n_{\textbf{act-params}}}$} &{\bf ARC} &{\bf MMLU} &{\bf HellaSwag} &{\bf Belebele} &{\bf Flores} &{\bf Avg.} \\
\midrule
\bf{Expanded Languages} &&&&&& \\
\midrule
Qwen1.5-1.8B &1.8B &1.8B &23.13 &30.97 &29.15 &33.15 &55.40 &34.36\\
\hdashline
LLaMA-Pro (+6 Layers) &2.1B &2.1B &24.89 &34.05 &33.03 &30.55 &79.82 &40.47\\
LLaMA-Pro (+12 Layers) &2.4B &2.4B  &24.35 &34.02 &33.85 &31.52 &81.76 &41.10\\
LLaMA-Pro (+24 Layers) &3B &3B &24.86 &33.68 &33.09 &27.11 &81.63 &40.07\\
\hdashline
LoRAMoE (Rank=32) &2B &2B &25.95 &33.64 &35.27 &31.30 &78.67 &40.97\\
LoRAMoE (Rank=180) &2.6B &2.6B &26.63 &34.17 &37.17 &32.81 &81.09 &42.37\\
\hdashline
MoE-LPR (2 Experts) &2.6B &2.6B &27.54 &34.36 &39.17 &38.26 &80.92 &44.05\\
MoE-LPR (4 Experts) &4.2B &2.6B &28.17 &35.24 &40.56 &39.44 &82.02 &45.09\\
MoE-LPR (6 Experts) &5.8B &2.6B &28.43 &34.10 &41.06 &39.93 &81.83 &45.07\\
MoE-LPR (8 Expert)s &7.4B &2.6B &29.03 &34.43 &41.67 &40.00 &81.91 &45.41\\
\midrule
\bf{Original Languages} &&&&&& \\
\midrule
Qwen1.5-1.8B &1.8B &1.8B &33.48 &47.55 &49.82 &56.52 &82.50 &53.97\\
\hdashline
LLaMA-Pro (+6 Layers) &2.1B &2.1B &32.71 &45.60 &48.76 &52.04 &82.10 &52.24\\
LLaMA-Pro (+12 Layers) &2.4B &2.4B  &31.77 &44.06 &48.36 &48.78 &81.97 &50.99\\
LLaMA-Pro (+24 Layers) &3B &3B &30.09 &40.02 &46.94 &40.63 &81.23 &47.78\\
\hdashline
LoRAMoE (Rank=32) &2B &2B &31.43 &43.63 &47.12 &44.63 &81.25 &49.61\\
LoRAMoE (Rank=180) &2.6B &2.6B &32.43 &45.41 &48.61 &47.74 &82.03 &51.24\\
\hdashline
MoE-LPR (2 Experts) &2.6B &2.6B &32.40 &43.60 &48.72 &49.81 &82.26 &51.36\\
MoE-LPR (4 Experts) &4.2B &2.6B &32.54 &44.75 &49.00 &50.67 &81.77 &51.75\\
MoE-LPR (6 Experts) &5.8B &2.6B &32.71 &44.62 &49.12 &51.81 &82.36 &52.12\\
MoE-LPR (8 Expert)s &7.4B &2.6B &32.20 &44.60 &48.83 &50.44 &82.18 &51.65\\
\midrule
\bottomrule
\end{tabular}}
\caption{Evaluation results on benchmarks for MoE-LPR with different numbers of experts, LLaMA-Pro with different extending layers and LoRAMoE with different ranks.}
\label{tab:scale}
\end{table*}

\end{document}